\documentclass[10pt,twocolumn,letterpaper]{article}

\usepackage{iccv}
\usepackage{times}
\usepackage{epsfig}
\usepackage{graphicx}
\usepackage{amsmath}
\usepackage{amssymb}
\usepackage{multirow}
\usepackage{booktabs}
\usepackage{hhline}
\usepackage{xcolor}
\usepackage{balance}
\usepackage[pagebackref=true,breaklinks=true,letterpaper=true,colorlinks,bookmarks=false]{hyperref}
\usepackage{flushend}

\usepackage{color}

\usepackage[breaklinks=true,bookmarks=false]{hyperref}

\iccvfinalcopy % *** Uncomment this line for the final submission

\def\iccvPaperID{****} % *** Enter the ICCV Paper ID here

\definecolor{myred}{RGB}{255, 0, 0}
\definecolor{mygreen}{RGB}{0, 176, 80}
\definecolor{myyellow}{RGB}{255, 192, 0}
\definecolor{myblue}{RGB}{0, 112, 192}

\begin{document}

%%%%%%%%% TITLE
%\title{Skip-Modal Generative Networks for Image-to-Speech Synthesis\thanks{Code: \url{https://github.com/yunyikristy/skipNet}}}
\title{Unpaired Image-to-Speech Synthesis with Multimodal Information Bottleneck\thanks{Code: \url{https://github.com/yunyikristy/skipNet}}}
%\title{Skip-Modal Generation with Multimodal Information Bottleneck}
%\title{Skip-Modal Correspondence Learning for Unpaired Image-to-Speech Synthesis\thanks{Code: \url{https://github.com/yunyikristy/skipNet}}}

\author{Shuang Ma\\
SUNY Buffalo\\
Buffalo, NY\\
{\tt\small shuangma@buffalo.edu}
\and
Daniel McDuff\\
Microsoft\\
Redmond, WA\\
{\tt\small damcduff@microsoft.com}
\and
Yale Song \\
Microsoft\\
Redmond, WA\\
{\tt\small yalesong@microsoft.com}
}

\maketitle
%\thispagestyle{empty}

%%%%%%%%% ABSTRACT
\begin{abstract}
Deep generative models have led to significant advances in cross-modal generation such as text-to-image synthesis. Training these models typically requires paired data with direct correspondence between modalities. We introduce the novel problem of translating instances from one modality to another without paired data by leveraging an intermediate modality shared by the two other modalities. To demonstrate this, we take the problem of translating images to speech. In this case, one could leverage disjoint datasets with one shared modality, e.g.,  image-text pairs and text-speech pairs, with text as the shared modality. We call this problem ``skip-modal generation'' because the shared modality is skipped during the generation process. We propose a multimodal information bottleneck approach that learns the correspondence between modalities from unpaired data (image and speech) by leveraging the shared modality (text). We address fundamental challenges of skip-modal generation: 1) learning multimodal representations using a single model, 2) bridging the domain gap between two unrelated datasets, and 3) learning the correspondence between modalities from unpaired data. We show qualitative results on image-to-speech synthesis; this is the first time such results have been reported in the literature. We also show that our approach improves performance on traditional cross-modal generation, suggesting that it improves data efficiency in solving individual tasks.
\end{abstract}

\begin{figure}[t]
    \centering
    \includegraphics[width=\linewidth]{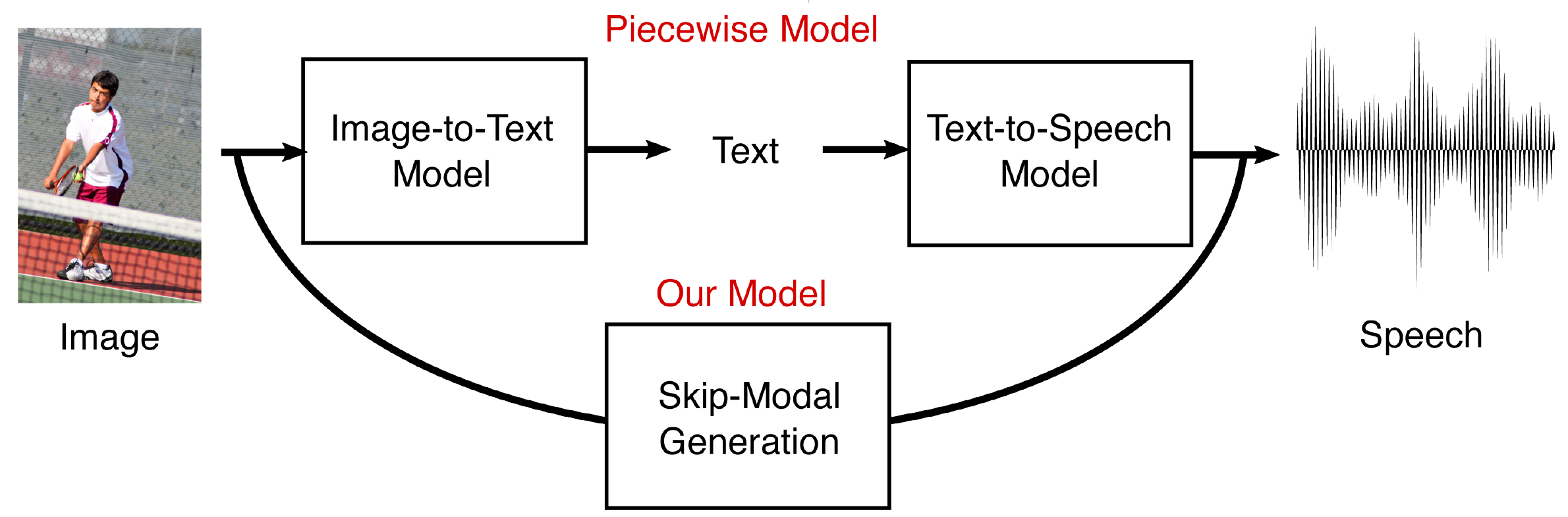}
    \caption{Cross-modal generation typically requires paired data with direct correspondence between modalities. However, this data is not always available (e.g., image-to-speech), in which case generation could be done by bridging two existing datasets via an intermediate modality (text). We propose an approach that directly generates outputs by learning multimodal correspondences from unpaired data provided by multiple disjoint datasets.} %learns multimodal representations using a single model, to bridge the domain gap between unrelated datasets, and learn the correspondence between modalities from unpaired data
\label{fig:concept}
%\vspace{-3mm}
\end{figure}

\vspace{-5mm} 
%%%%%%%%%%%%%%%%%%%%%%%%%%%%%%%%%%%%%%%%%%%%%%%%%%%%%%%%%
\section{Introduction}
%%%%%%%%%%%%%%%%%%%%%%%%%%%%%%%%%%%%%%%%%%%%%%%%%%%%%%%%%
Recent advances in deep generative models have shown impressive results across many cross-modal generation tasks, including text-to-image~\cite{reed2016generative}, text-to-speech~\cite{ma2019tts}, image-to-video~\cite{mathieu2016deep}, video-to-sound~\cite{zhou2018visual} synthesis. Training these models typically requires a large amount of paired samples with direct correspondence between instances from the different modalities, which limits their applicability to new (``unseen") modalities. Some attempts have been made to eliminate such constraint in the context of image-to-image cross-\textit{domain} translation, training a network on unpaired examples with the cycle consistency constraint~\cite{cycleGAN,Bicycle-GAN,stargan}. However, those methods generally assume that two domains come from the same modality, e.g., images of horses and zebras; as we show later, these methods tend to fail in a cross-\textit{modal} scenario (such as image-to-speech) where the assumption no longer holds. 

In this work, we aim to learn a mapping from one modality to another without using paired samples. Our main idea is to leverage readily available datasets that do not directly provide paired samples of the two modalities we are interested in, but have ``skip'' correspondence between the two desired modalities via a shared one. For example, for image-to-speech synthesis we may leverage two existing datasets with image-text and text-speech pairs, where text serves as the shared modality. A naive solution to this would be training two networks separately, each solving either of the tasks with the paired data, and running them sequentially, e.g., given an image, generate text, and use it to generate speech. However, this approach is not trainable end-to-end and suffers from several issues such as domain discrepancy and information loss between the two models.

We introduce a new task \textit{skip-modal generation} that aims to translate one modality to another by ``skipping'' an intermediate modality shared by two different datasets. There are several reasons why this is an interesting problem to solve. From a practical standpoint, leveraging readily available datasets for solving new tasks allows for new applications. Also, training a single model with multiple datasets could potentially improve data efficiency, improving performance on the tasks each dataset was originally designed for; later, we empirically show this is indeed the case with our proposed model. From a theoretical standpoint, an ability to translate across multiple modalities may suggest that the model is one step closer to finding a unified abstract representation of different sensory inputs~\cite{pietrini2004beyond,giard1999auditory}. Achieving this means information from one can be translated into any of the other modalities. Our experiments show our proposed approach can translate instances across different combinations of image, text, and speech modalities.   

%\sm{We find that, by reaching the skip-model generation goal, our model learns unified representations from multi-modality data jointly. The elaborate designed Memory Fusion Module serves as an external memory which stores all information from multiple modalities and multiple domains in a compact latent space. It is learned during the training process without any supervision. Our proposed multi-task driven training objective makes the our model becomes a generative one which can be used in a large scale of down-stream applications. }
%Our inspiration is partially derived from neuroscience research that has revealed that the brain forms unified abstract representations from cross-sensory modalities~\cite{pietrini2004beyond,giard1999auditory}. For instance, the same mirror neurons are shown to fire when primates observe actions and hear their associated sounds~\cite{kohler2002hearing,giard1999auditory}; visual and tactile recognition uses similar processes~\cite{pietrini2004beyond}. While there is much we do not understand about the nature of how the brain processes multi-sensory signals, we are motivated by this property in the design of our model.

We focus on addressing three key challenges in skip-modal generation: learning to represent multimodal data in a uniform manner, resolving multi-dataset domain discrepancies, and learning the correspondence from unpaired data. To this end, we propose a novel generative model trainable on multiple disjoint datasets in an end-to-end fashion. Our model consists of modality-specific encoders/decoders and a multimodal information bottleneck (MIB) that learns how to represent different modalities in a shared latent space. The MIB transforms each modality-specific encoder output into the shared modality space (e.g., text) and further processes it through a memory network that serves as an information bottleneck~\cite{tishby1999information}. This helps us obtain unified abstract representations of multiple modalities, capturing \textit{``the most meaningful and relevant information''}~\cite{tishby1999information} regardless of modalities or datasets. We train our model by solving two cross-modal generation tasks through the shared modality, enabling the model to learn multimodal correspondence.

We evaluate our approach on image-to-speech synthesis using two existing datasets -- the COCO~\cite{MSCOCO} dataset that provides image-text pairs, and an in-house text-to-speech (TTS) dataset that provides text-speech pairs -- and demonstrate a superior performance over current baselines. To the best of our knowledge, this is the first time image-to-speech synthesis results have been reported. We also evaluate our approach on each of the cross-modal generation tasks the datasets were originally collected for, and show that we outperform previous state-of-the-art methods on each task, suggesting our method also improves data efficiency.

To summarize our contributions, we: (1) introduce skip-modal generation as a new task in multimodal representation learning; (2) propose an approach that learns the correspondence between modalities from unpaired data; (3) report realistic image-to-speech synthesis results, which has not been reported in the literature before; (4) show our model improves data efficiency, outperforming previous results on cross-modal generation tasks.

\begin{figure*}[t]
\begin{center}
    \includegraphics[width=0.93\linewidth]{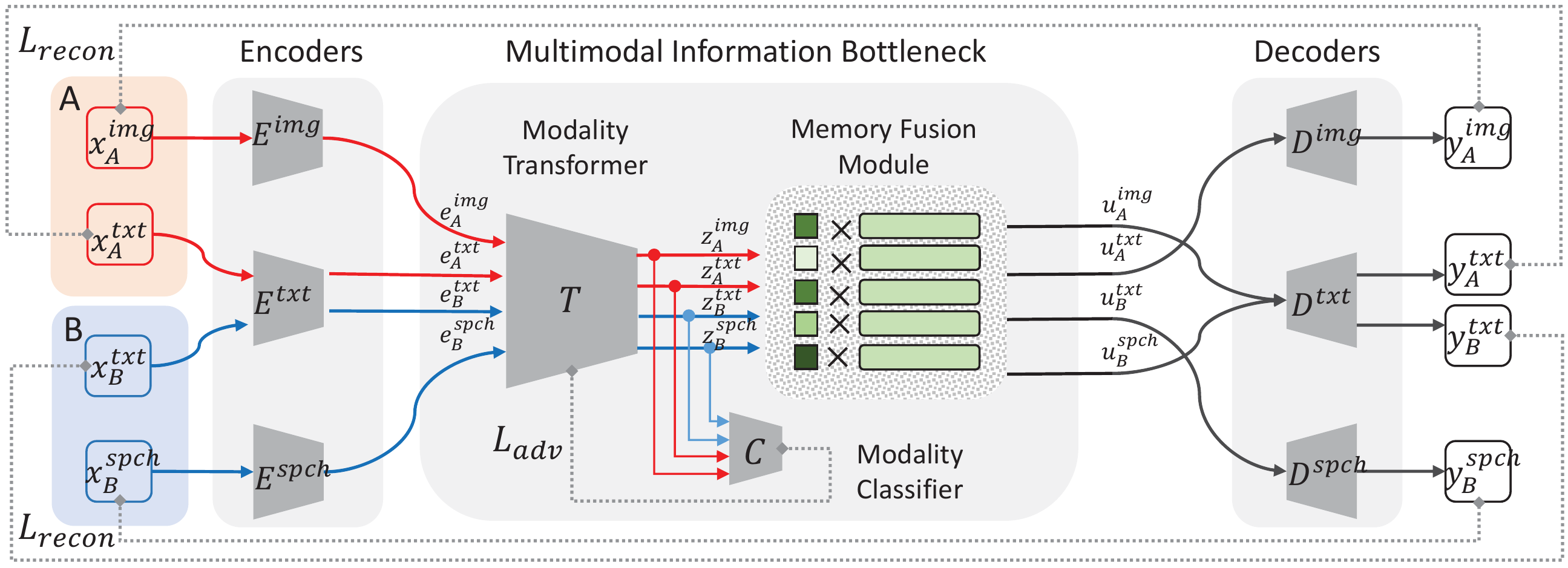}
\end{center}
   \caption{Our model consists of modality-specific encoders $\mathbf{E}^{(\cdot)}$ and decoders $\mathbf{D}^{(\cdot)}$, and a multimodal information bottleneck that learns to represent different modalities in a shared embedding space. We train the model end-to-end using two disjoint datasets involving image, text, and speech (A and B) by solving cross-modal reconstruction tasks. The modality classifier $\mathbf{C}$ is used only during training.}
\label{fig:pipeline}
%\vspace{-3mm}
\end{figure*}

%%%%%%%%%%%%%%%%%%%%%%%%%%%%%%%%%%%%%%%%%%%%%%%%%%%%%%%%%
\section{Related Work}
%%%%%%%%%%%%%%%%%%%%%%%%%%%%%%%%%%%%%%%%%%%%%%%%%%%%%%%%%
%\vspace{-2mm}
\textbf{Cross-Modal Synthesis}: There has been much progress in cross-modal synthesis involving language, vision, and sound. For vision and language, image-to-text synthesis (image captioning) has been a popular task, where attention mechanisms have shown particularly strong results~\cite{show&tell,show-attend-and-tell,caption-att,style-caption,caption-att-1,caption-att-2}. In text-to-image synthesis, most existing methods are based on deep generative models~\cite{GANs,kingma2014auto}. Reed~\etal~\cite{reed2016generative} and Zhang~\etal~\cite{stackgan} were some of the first to show promising results. Further improvements have been reported using attention mechanisms~\cite{AttnGAN, DA-GAN}. 
For language and sound, speech-to-text (ASR) is perhaps the most mature topic of research, and great advances have been made with deep learning~\cite{hinton2012deep}. Text-to-speech synthesis using deep neural networks has gained much attention recently, with methods such as WaveNet \cite{WaveNet}, DeepVoice \cite{DeepVoice1, NIPS2017:DeepVoice2, ICLR2018:DeepVoice3}, VoiceLoop \cite{voiceloop, fitting-new-speakers}, Char2Wav \cite{Char2Wav}, and Tacotron \cite{tacotron, Tacotron-gst}. 
Our work is distinct from all existing lines of research in cross-modal synthesis in that we do not require paired samples to train a model. Instead, we leverage a shared modality between different datasets to learn the skip-correspondence between modalities where no paired data is available.

\textbf{Cross-Domain Synthesis:} 
Cross-domain within-modality synthesis has also been a topic of extensive study. Pix2pix~\cite{pix2pix} was the first attempt at translating across different image domains by training on paired data (e.g., sketches to photos). Since then, numerous methods have tackled the problem from an unsupervised learning perspective, eliminating the need for paired data~\cite{cycleGAN,UNIT, DA-GAN, distance-GAN, DiscoGAN}. Methods based on cycle consistency~\cite{cycleGAN} have been particularly effective in this problem space. Unfortunately, cross-domain synthesis methods tend to fail on cross-modal scenarios because of the larger domain gap between different modalities. We empirically validate this in our experiments.
Instead of using the cycle consistency loss, Lior~\etal~\cite{UNIT} translate between human faces and emojis. They leverage the fact that a face has a rigid low-dimensional structure (e.g., facial landmarks), and use a pretrained human face classifier to obtain effective representations of both human faces and emojis. Unlike their approach, in this work we make no assumption about the types of data.

%%%%%%%%%%%%%%%%%%%%%%%%%%%%%%%%%%%%%%%%%%%%%%%%%%%%%%%%%
\section{Approach}
%%%%%%%%%%%%%%%%%%%%%%%%%%%%%%%%%%%%%%%%%%%%%%%%%%%%%%%%%
%\vspace{-2mm}
Given two cross-modal datasets with one shared modality -- e.g., a text-image dataset $\mathcal{A}=\{(\mathbf{x}^{txt}_{A,i}, \mathbf{x}^{img}_{A,i})\}_{i=1}^{N}$ and a text-speech dataset $\mathcal{B}=\{(\mathbf{x}^{txt}_{B,i}, \mathbf{x}^{spch}_{B,i})\}_{i=1}^{M}$, with text as a shared modality -- our goal is to learn a network that can model data from all three modalities. We design our network with modality-specific encoders and decoders $\mathbf{E}^{j} \mbox{ and } \mathbf{D}^{j}$, respectively, with $ j=\{\mbox{text, image, speech}\}$. Note that the definition of our model is agnostic to modalities; the encoders/decoders can be swapped out for different modalities depending on the dataset and application.  % For simplicity, we drop the subscript indices unless the distinction is necessary. 

Our main technical contribution is the multimodal information bottleneck (MIB), which consists of a modality transformer $\mathbf{T}$ and a memory fusion module $\mathbf{M}$ (see Figure~\ref{fig:pipeline}); the modality classifier $\mathbf{C}$ is used only during training. The $\mathbf{T}$ transforms output from different encoders into the shared modality space (text); the $\mathbf{M}$ acts as an information bottleneck~\cite{tishby1999information} and further processes the signals to produce compact, unified abstract representations. We use the output to generate an instance in different modalities.

\begin{figure*}[t]
\begin{center}
    \includegraphics[width=1.0\linewidth]{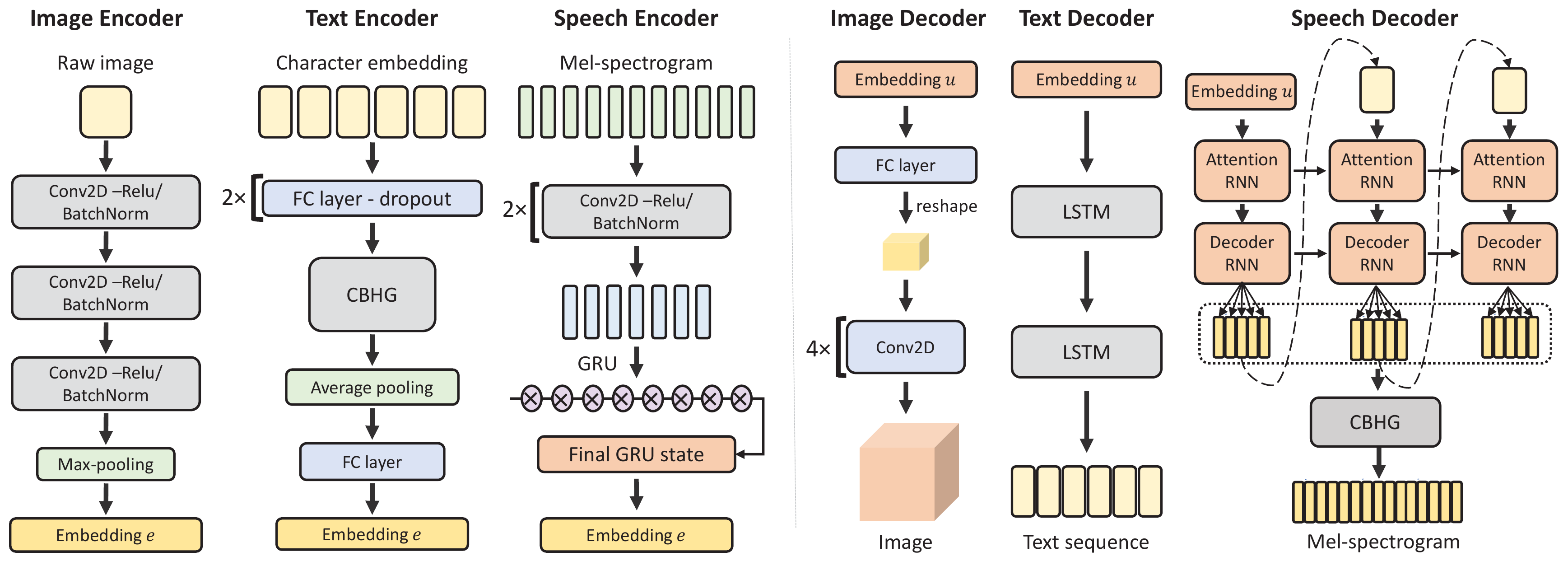}
\end{center}
   \caption{\textbf{Architectures of the modality-specific encoders and decoders.} We provide parameter settings in the supplementary material.}
\label{fig:encdec_arch}
%\vspace{-3mm}
\end{figure*}

%--------------------------------------------------------
\subsection{Modality-Specific Encoders}

\textbf{Image encoder}: We feed images to a three-layer CNN and perform max-pooling to obtain the output $\mathbf{e}^{img} \in \mathbb{R}^{512}$.
% with 64, 128, 256, 512 filters for each layer, 

\textbf{Text encoder}: We process text into a sequence of 128-D character-level embeddings via a 66-symbol trainable look-up table. We then feed each of the embeddings into two fully-connected (FC) layers. The output sequence is fed into the CBHG~\cite{tacotron} to obtain a sequence of 128-D embeddings; we use the original parameter settings of~\cite{tacotron}. Finally, we apply average pooling over the sequence and feed it into one FC layer with 512 units to obtain the output $\mathbf{e}^{txt} \in \mathbb{R}^{512}$.
%  with 256 and 128 units

\textbf{Speech encoder}: We extract mel-spectrograms, a time-frequency representation of sound, from audio waveforms using 80 frequency bands. We treat this as a single-channel image of dimension $t$-by-80, where $t$ represents the time. We feed it into a two-layer fully convolutional network and further process it using a GRU~\cite{cho2014learning} with 512 units, feeding in a 5-by-80 chunk at a time. We take the last state of the GRU as the output $\mathbf{e}^{spch} \in \mathbb{R}^{512}$.  

%--------------------------------------------------------
\subsection{Multimodal Information Bottleneck}

Neuroscientists have developed theories that the brain forms unified representations of multimodal signals~\cite{giard1999auditory,pietrini2004beyond}. Modelling this computationally is very challenging because information contained in different modalities are often not directly comparable. The mapping of instances between modalities are not bijective, nor injective, nor surjective. This is especially true between text and image/speech; a sentence ``There is a little blue bird'' can map to images depicting different shapes and poses of a bird, or to speech signals with different intonation, tone, stress, and rhythm. Conversely, certain imagery and sounds are indescribable. 

To tackle our problem of modeling multimodal data despite these challenges, we focus on how \textit{structured} and \textit{compact} textual representations are; image and audio contain richer information with considerably higher degrees of variability than text. Thus, we use text as a \textit{conduit} to learn the correspondence between image and speech. This has an effect of an information bottleneck~\cite{tishby1999information}, which limits the flow of certain modality-specific information and helps the model learn to align image and speech from unpaired data.

\textbf{Modality transformer}: We start by transforming instances from image and speech modalities into a shared latent space induced by the text modality. The modality transformer $\mathbf{T}$ is a three-layer residual network that maps embeddings of each modality $\mathbf{e}^{j}$ to $\mathbf{z}^{j}\in\mathbb{R}^{256}$. 

To ensure the desired transformation is performed, we use an adversarial objective that encourages $\mathbf{z}^{j}$ to be indistinguishable from each other with respect to the text modality. To this end, we design a modality classifier $\mathbf{C}$ with two FC  layers and a 3-way softmax classifier representing the three modalities. We then define an adversarial loss as
\begin{equation}
\label{eq:loss_adv}
\mathcal{L}_{adv} = \mathop {\min }\limits_{\mathbf{T}} \mathop {\max }\limits_{\mathbf{C}} {\mathcal{L}_{\mathbf{T}}} + {\mathcal{L}_{\mathbf{C}}}
\end{equation} 
where the mini-max game is defined with two terms
\begin{align}
\mathcal{L}_{\mathbf{T}} = 
& - \mathbb{E} \big[\log \mathbf{C}(\mathbf{T}(\mathbf{e}^{img}_A))_{txt}\big] 
  - \mathbb{E} \big[\log \mathbf{C}(\mathbf{T}(\mathbf{e}^{spch}_B))_{txt}\big] \nonumber \\
\mathcal{L}_{\mathbf{C}} = 
& - \mathbb{E} \big[\log \mathbf{C}(\mathbf{z}^{img}_A)_{img}\big] 
  - \mathbb{E} \big[\log \mathbf{C}(\mathbf{z}^{spch}_B)_{spch}\big] \nonumber \\
& - \mathbb{E} \big[\log \mathbf{C}(\mathbf{z}^{txt}_A)_{txt}\big]
  - \mathbb{E} \big[\log \mathbf{C}(\mathbf{z}^{txt}_B)_{txt}\big] \nonumber
\end{align}
where $\mathbf{C}(\cdot)_{j}$ means we take the value from the corresponding category. To make an analogy to GAN training~\cite{GANs}, $\mathbf{C}$ acts as a modality discriminator and $\mathbf{T}$ tries to fool $\mathbf{C}$ into believing that all $\mathbf{z}^{j}$ are from the text modality. In practice, we add the gradient reversal layer~\cite{ganin2015unsupervised} to train our model without having to alternative between min-max objectives.

\textbf{Memory fusion module}: Next, we extract the uniform abstract representation $\mathbf{u}^{j}$ which has the most relevant information shared between paired modalities. A principled way to achieve this is through the information bottleneck (IB) approach~\cite{tishby1999information}, which seeks a coding mechanism that maximally preserves information in the input signal when represented using a set of external variables.

The design of our memory fusion module is partly inspired by memory networks~\cite{weston2015memory} and multi-head self-attention~\cite{vaswani2017attention}. In a nutshell, we define an external memory $M$ that stores basis vectors representing modality-agnostic ``abstract concepts,'' which is shared by all the modalities involved. The model reads from the memory during the forward pass, and writes to it during back-propagation. We use multi-head self-attention~\cite{vaswani2017attention} as our coding mechanism, encoding $\mathbf{z}^{j}$ into $\mathbf{u}^{j}$ with respect to the shared $M$. 

Formally, we define an external memory $M \in \mathbb{R}^{n_k \times d_k}$, where $n_k$ is the number of basis vectors and $d_k$ is the dimension of each basis vector. We also define an intermediate variable $K \in \mathbb{R}^{n_k \times d_k}$ which we use with $M$ to form the ``$\langle$key, value$\rangle$ pairs'' for the multi-head self-attention ($K$ is the key, $M$ is the value). We compute $K$ by convolving $M$ with 256 1D kernels of size one. Finally, we compute $\mathbf{u}^{j}$ as a linear combination of basis vectors in $M$ with weights given by the scaled dot-product attention~\cite{vaswani2017attention},
\begin{equation}
\label{eq:attn}
\mathbf{u}^{j} = \mbox{softmax} \left( \mathbf{z}^{j} K^T / \sqrt{d_k} \right) M
\end{equation}
Intuitively, $\mathbf{z}^{j}$ serves as a query to search the relevant keys to determine \textit{where to read} from the memory. The scaled dot-product inside the softmax can be understood as a compatibility function between a query and the keys, which gives us attention scores for attending to different parts of the memory. We use multi-head self-attention with four parallel heads to make the module jointly attend to information from different subspaces at different positions. 

\begin{figure}[t]
\begin{center}
    \includegraphics[width=1.0\linewidth]{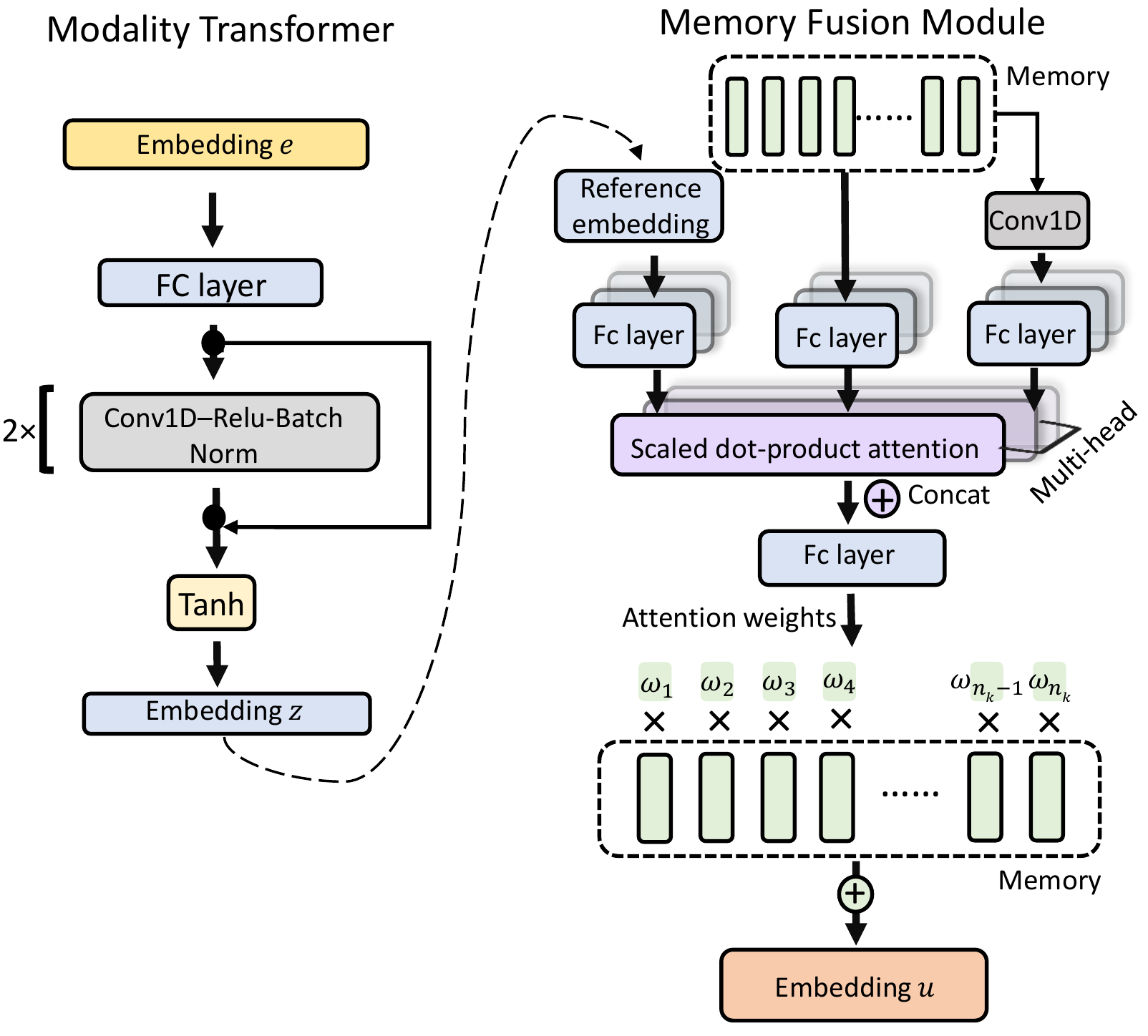}
\end{center}
   \caption{\textbf{Architecture of multimodal information bottleneck.}}
\label{fig:ucb_arch}
%\vspace{-3mm}
\end{figure}

\textbf{Training of the memory fusion module:} Enabling the desired information bottleneck effect requires a careful design of the learning objective. One popular way is to impose an autoencoder-type reconstruction loss for each modality. In our scenario, this corresponds to, e.g., $\mathbf{u}^{img}$ reconstructing $\mathbf{x}^{img}$. While this would help the network learn to bottleneck superfluous information present within each modality, it would miss out on the opportunity to learn \textit{cross-modal correspondences}. Most crucially, this will prevent the network from learning \textit{modality-agnostic} representations, which are important for skip-modal generation, and instead learn redundant concepts presented by different modalities.

Therefore, we solve the cross-modal generation tasks as provided by two paired datasets. Specifically, we aim to reconstruct $\mathbf{x}^{j}_{A}$ from $\mathbf{u}^{j}_{A}$ and  $\mathbf{x}^{j}_{B}$ from $\mathbf{u}^{j}_{B}$ in a cross-modal fashion. We define our loss as
\begin{equation}
\label{eq:recon}
    \mathcal{L}_{recon} = \mathcal{L}^{img}_{A} + \mathcal{L}^{spch}_{B} + \mathcal{L}^{txt}_{A,B}
\end{equation}
We use the $l_1$ loss for both image and speech modalities:
\begin{align}
    \mathcal{L}^{img}_{A} = \frac{1}{N}\sum_{i=1}^{N} 
        \| \mathbf{x}^{img}_{A,i} - \mathbf{D}^{img}(\mathbf{u}^{txt}_{A,i})\|_{1} \\
    \mathcal{L}^{spch}_{B} = \frac{1}{M}\sum_{i=1}^{M} 
        \| \mathbf{x}^{spch}_{B,i} - \mathbf{D}^{spch}(\mathbf{u}^{txt}_{B,i})\|_{1} 
\end{align}
For text modality we use the cross-entropy loss:
\begin{align}
    \mathcal{L}^{txt}_{A,B} = 
    & - \frac{1}{N}\sum_{i=1}^{N} 
        \mbox{CE}\left(\mathbf{x}^{txt}_{A,i}, 
            \mathbf{D}^{txt}(\mathbf{u}^{img}_{A,i})\right) \nonumber \\
    & - \frac{1}{M}\sum_{i=1}^{M} 
        \mbox{CE}\left(\mathbf{x}^{txt}_{B,i}, 
            \mathbf{D}^{txt}(\mathbf{u}^{spch}_{B,i})\right) 
\end{align}
where we compare two sentences character-by-character according to 66 symbol categories. Note that the computation of $\mathcal{L}^{txt}_{A,B}$ depends on both $\mathcal{A}$ and $\mathcal{B}$, and the text decoder must serve a dual-purpose as an image-to-text generator and a speech-to-text generator. This allows our network to learn the skip-modal correspondence between image and speech. It also maximizes the information bottleneck effect in our memory fusion module because the external memory is conditioned on all three combinations of the modalities.

\textbf{Interpretation of the multimodal information bottleneck:} The two components in the MIB compensate each other with related yet different objectives. The modality transformer ``drags'' any given modality-specific embedding $\mathbf{e}^{j}$ into a shared latent subspace induced by the text modality. This helps us further process the signals in a more stable manner; otherwise the memory fusion module must deal with signals coming from three different spaces, which may have different statistical properties. 

The memory fusion module then encourages $\mathbf{u}^{j}$ to contain the most relevant correspondence information between modalities. We share the external memory to encode embeddings from different modalities. Trained with our cross-modal reconstruction objectives, the use of a shared memory provides a strong ``bottleneck'' effect so that (1) it suppresses modality-specific information that does not contribute to cross-modal generation, and (2) it focuses on finding a highly-structured latent multimodal space. This allows us to obtain compact representations of the multimodal data. In Section~\ref{sec:experiments}, we show this improves not only the generalization ability for skip-modal generation, but also the data efficiency for each individual cross-modal generation task.

\begin{figure*}
    \centering
    \includegraphics[width=\linewidth]{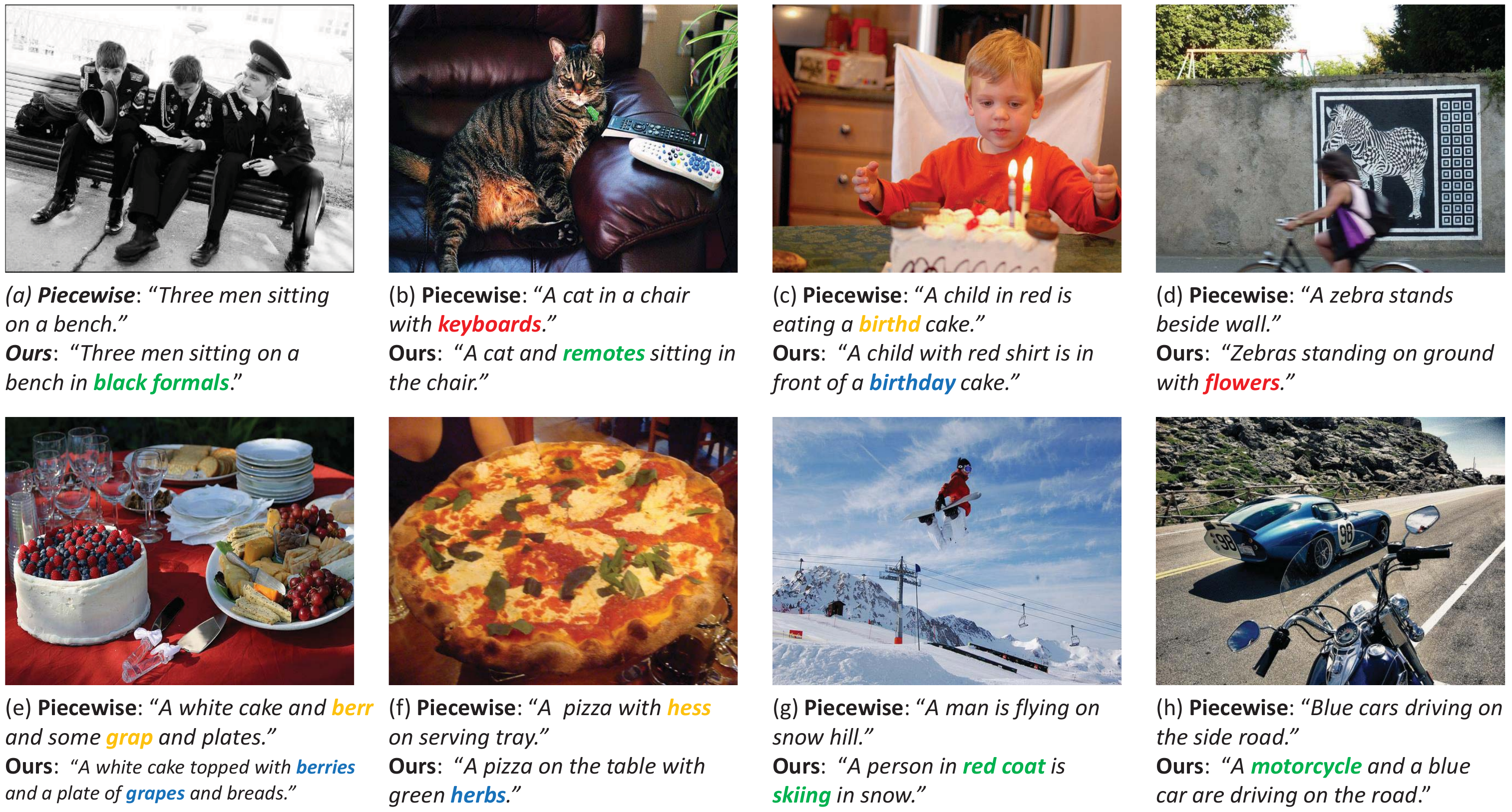}
    \caption{\textbf{Image-to-speech synthesis results.} For the purpose of presentation, we manually transcribed audio results. \textcolor{myred}{Red}: incorrect word predictions, \textcolor{mygreen}{green}: correct/more fine-grained word predictions compared with the baseline, \textcolor{myyellow}{yellow}: incorrect word pronunciation, and \textcolor{myblue}{blue}: correct/better word pronunciation compared with the baseline. Audio samples are available at \url{https://bit.ly/2U7741S}}
    \label{fig:examples}
    %\vspace{-3mm}
\end{figure*}

%--------------------------------------------------------
\subsection{Modality-Specific Decoders}
\textbf{Image decoder}: We feed $\mathbf{u}^{j}$ into one FC layer with 1024 units and reshape the output to be in $R^{4 \times 4 \times 64}$. We then upsample it with four deconvolutional layers to generate an image of size 128 $\times$ 128 pixels. During training, we feed $\mathbf{u}^{txt}$ to the decoder for cross-modal generation.

\textbf{Text decoder}: We use a two-layer LSTM as our text decoder. After initializing it with $\mathbf{u}^{j}$, we unroll it to generate a sentence until we get the end-of-sentence token. During training, we feed either $\mathbf{u}^{img}$ or $\mathbf{u}^{spch}$ to the decoder.

\textbf{Speech decoder}: We use the attention-based decoder of \cite{decoder-speech} that contains an attention RNN (a two-layer residual GRU with 256 cells) and a decoder RNN (a single-layer GRU with 256 cells). We initialize both RNNs with $\mathbf{u}^{j}$ and unroll them to generate a series of $t$-by-80 mel-spectrogram chunks. At each step, we predict multiple, non-overlapping chunks, which has been shown to speed up the convergence~\cite{tacotron}. We convert the predicted mel-spectrogram into an audio waveform using the Griffin-Lim algorithm~\cite{griffin-lim}. During training, we feed $\mathbf{u}^{txt}$ to the decoder, while at inference time we feed $\mathbf{u}^{img}$ for skip-modal generation. %, where a recurrent layer provides an attention query to the decoder at each step. It

%--------------------------------------------------------
\subsection{Learning Objective and Optimization}
We train our model by minimizing a loss function
\begin{equation}
    \mathcal{L} = \mathcal{L}_{recon} + \alpha \mathcal{L}_{adv}
\end{equation}
where we set $\alpha=0.1$ in our experiments. We train the whole network end-to-end from scratch using the ADAM optimizer~\cite{kingma2014adam} with an initial learning rate of 0.002. We train our model for 100 epochs using a batch size of eight.% on four NVIDIA M40 GPUs.

%%%%%%%%%%%%%%%%%%%%%%%%%%%%%%%%%%%%%%%%%%%%%%%%%%%%%%%%%
\section{Experiments}
\label{sec:experiments}
%%%%%%%%%%%%%%%%%%%%%%%%%%%%%%%%%%%%%%%%%%%%%%%%%%%%%%%%%
We evaluate our proposed approach from two perspectives: 1) image-to-speech synthesis; 2) the effectiveness of multimodal modeling. We train our model on two datasets: COCO \cite{MSCOCO} that contains image-text samples, and an in-house dataset EMT-4 that contains 22,377 American-English audio-text samples, with a total of 24 hours. All the audio samples are read by a single female speaker.

%--------------------------------------------------------
\subsection{Skip-Modal Generation}
\label{sec:vs.img-speech}
We validate skip-modal generation on image-to-speech synthesis both qualitatively and quantitatively, comparing ours with two baselines: the piecewise approach and CycleGAN~\cite{cycleGAN}. The piecewise approach uses two individual models~\cite{caption-att} sequentially, e.g., image-to-text followed by text-to-speech. CycleGAN~\cite{cycleGAN} was originally proposed for image-to-image translation from unpaired data. To see how the model generalize to the cross-modal case, we train it directly on images and audio samples from both datasets. For a fair comparison, we design both baselines using the same encoder-decoder architectures as ours, and train them end-to-end from scratch using the same dataset.

\textbf{Qualitative evaluation}.
We had seven human judges evaluate the generated speech from our skip-generation model and the two baselines. Twenty speech samples were generated using the models, leading to 140 independent evaluations.  The judges were shown the source image and listened to the speech. They were asked to select the audio sample that had the most accurate content and the sample with speech that was closest to a human voice. They also selected the sample they felt had the highest overall quality. Examples of the samples can be found here: \url{https://bit.ly/2U7741S}
%\url{https://sites.google.com/view/image-to-speech-demo-page/home}.

On average 78.6\% (sd = 27.6\%) of the subjects picked ours for the highest quality content. Based on audio quality, 65.0\% (sd = 35.7\%) of the subjects picked ours as the highest quality. Based on overall quality, 74.3\% (sd = 33.9\%) of the subjects picked ours. In summary, our subjects picked ours three times more frequently than either of the other baselines based on all three quality metrics.

Figure~\ref{fig:examples} shows some of the samples used in our user study; we manually transcribed the synthesized audio results for the purposes of presentation. We analyze the results by focusing on two aspects: 1) does the speech sample correctly describe the content as shown in the image? 2) is the quality of pronunciation in the speech sample realistic? 

The piecewise approach sometimes incorrectly predicted words, e.g., in Fig.~\ref{fig:examples} (b) keyboards vs. remotes. We also see that our approach produces results with more fine-grained details, e.g., (g) flying vs. skiing, (h) motorcycle is missed by the baseline. These suggest that our approach is superior to the baseline in terms of modeling multimodal data.  

One limitation of the piecewise approach is the inability to deal with the \textit{domain gap} between datasets, e.g., certain concepts appear in one dataset but not in the other. This is indeed our case: the vocabularies of the two datasets overlap by only 26\% (COCO has 15,200 words and EMT-4 has 17,946 words; 6,874 words overlap). This domain gap issue is reflected in our results: (e) the pronunciation of 'berries' and 'grapes' are incorrect in the baseline result, and similarly for (c) and (f). These words (berries, grapes, birthday, herb) do not appear in the EMT-4 dataset, which means the text-to-speech model must perform zero-shot synthesis. This is reflected in Fig.~\ref{fig:examples} (c, e, f) - see the yellow words. Our results show superior quality on those out-of-vocabulary words despite being trained on the same datasets. To quantify the word expressivity of our model, we analyzed the vocabulary size of the synthesized speech using ASR~\cite{WaveNet}. Our model produced a vocabulary of 2,761 unique words, while the piecewise baseline produced 1,959 unique words; this is 802 more words, a 40$\%$ increase over the baseline. %We believe this result further supports the benefit of multimodal joint learning.
%This suggests the superiority of our model is in the modeling of the multimodal data when compared to the baseline.  

\begin{figure*}
    \centering
    \includegraphics[width=\linewidth]{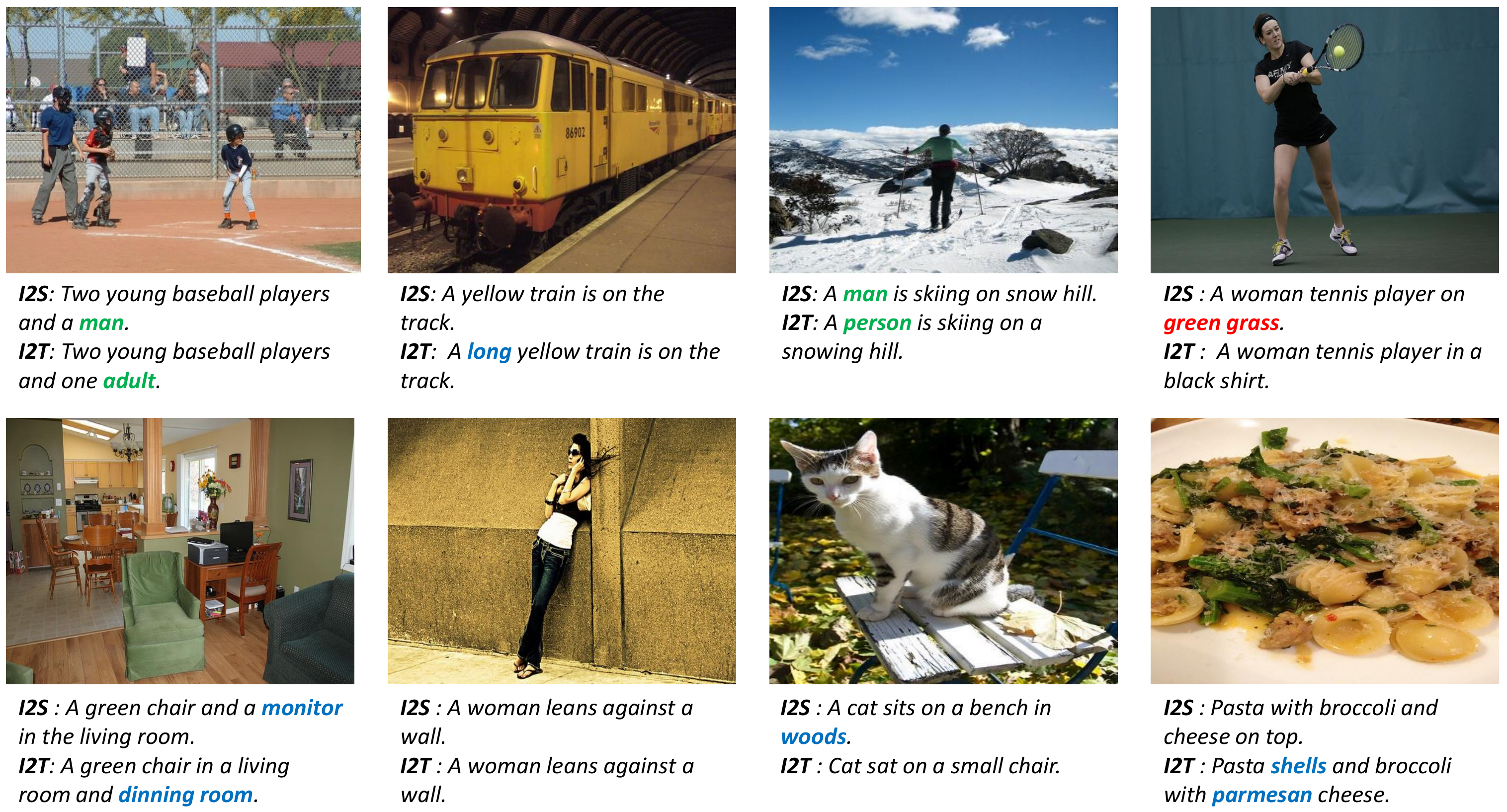}
    \caption{\textbf{Image-to-speech (I2S) and image-to-text (I2T) synthesis results}. I2S results are manually transcribed for presentation.}
    \label{fig:i2st}
    \vspace{-3mm}
\end{figure*}

Finally, we show additional results in Figure~\ref{fig:i2st} where we synthesize both speech and text from the same image as an input (speech results are manually transcribed). We see that the text and speech results are semantically very similar in that they describe the same content. This, together with other results above, suggests the model has learned to extract a unified abstract representation of multimodal data because different decoders can reliably synthesize samples that contain similar content -- despite the speech decoder having never seen the image embeddings during training.  

\textbf{Quantitative evaluation}.
To evaluate image-to-speech synthesis results quantitatively, we use a pretrained ASR model based on WaveNet \cite{WaveNet} and compare the text output with the ground-truth sentence corresponding to an input image from COCO. We report the results using the BLEU scores and the word error rate (WER). Table~\ref{tab:img2aud} shows our approach achieving the lowest WER with the highest BLEU scores (except for BLEU-4). 

\begin{table}
    \centering
    \small
    \begin{tabular}{rrrrrr}
    %\hline
    %& \multicolumn{5}{c}{Evaluation Metric} \\ \hline
    \toprule
               & B@1          & B@2            &B@3  &B@4        & WER           \\ \hline \hline
    CycleGAN \cite{show-attend-and-tell}  &26.2  &20.1    &11.3    &9.2           & 12.1   \\
    Piecewise \cite{caption-att}   &68.2                &51.9    &39.2     & \textbf{30.1} & 4.1             \\ \hline
    Ours            & \textbf{69.2}     & \textbf{52.1}   &\textbf{40.8}       & 29.9   &\textbf{3.9}      \\
    \bottomrule
    \end{tabular}
    \vspace{0.2cm}
    \caption{\textbf{Skip-modal generation results.} B@k are BLEU scores.}
    \label{tab:img2aud}
    %\vspace{-3mm}
\end{table}

%--------------------------------------------------------
\subsection{Cross-Modal Generation} 
%\vspace{-2mm}
\label{sec:vs.single-task}
To evaluate our approach in an objective manner, we turn to cross-modal generation where there exist state-of-the-art approaches and widely used metrics for each task. 

\begin{table}
    \centering
    \small
    % \begin{tabular}{rrrrrrr}
    \begin{tabular}{rp{0.65cm}p{0.65cm}p{0.65cm}p{0.65cm}p{0.9cm}p{0.7cm}}
    \toprule
               & B@1          & B@2         & B@3  & B@4    & CIDEr     & SPICE        \\ \hline \hline
    ATT \cite{caption-att}   &70.9                &53.7    &40.2     &30.4   & -- & --           \\
    SAT \cite{show-attend-and-tell}  &71.8  &50.4    &35.7    &25.0        & -- & --       \\
    RFNet \cite{Anderson2017up-down}  &76.4  & 60.4  &46.6   &35.8   &112.5   &20.5 \\
    UD \cite{RFNet}    &77.2  & --    & --     &36.2   &113.5   &20.3 \\\hline
    Ours  & 74.1   & 55.2  &41.1 &30.6  & --  & --  \\
    w/ \cite{RFNet} &78.9  &63.2  &48.1  & 37.0  &116.2  & 22.4 \\
    w/ \cite{Anderson2017up-down} & \textbf{79.8}     & \textbf{64.0}  & \textbf{48.9}  & \textbf{37.1}  &\textbf{117.8}  & \textbf{22.5} \\
    \bottomrule
    \end{tabular}
    \vspace{0.02cm}
    \caption{\textbf{Image-to-text generation results on COCO.}}
    \label{tab:eva_img2txt}
\end{table}

\begin{table}
    \centering
    \small
    \begin{tabular}{rrcrr} 
    \addlinespace[-\aboverulesep] 
    \cmidrule[\heavyrulewidth]{1-2} \cmidrule[\heavyrulewidth]{4-5}
    &  WER & \multirow{4}{*}{} & & WER  \\ %\cline{1-2}\cline{4-5} \morecmidrules \cline{1-2}\cline{4-5} 
    \hhline{==~==}
    Policy \cite{Zhou2018ImprovingES} & 5.42 & & DeepVoice3 \cite{deepvoice3} & 10.8\\ 
    DeepSpeech2 \cite{DBLP:DeepSpeech2}      & 5.15 & & Tacotron \cite{tacotron} & 10.6 \\ 
    GateConv \cite{GatedConvnet} & 4.80 & & Tacotron2 \cite{Tacotron2} & 10.5 \\
    Seq2Seq \cite{speech-recog-seq2seq}      & 4.01   & & GST \cite{Tacotron-gst} & \textbf{10.2} \\
    \cline{1-2}\cline{4-5} 
    Ours    & \textbf{3.88}  & & Ours & 10.5 \\
    \addlinespace[-\aboverulesep] 
    \cmidrule[\heavyrulewidth]{1-2} \cmidrule[\heavyrulewidth]{4-5}
    \end{tabular}
    \vspace{0.02cm}
    \caption{\textbf{Speech-to-text (left) and text-to-speech (right) results.}}
    \label{tab:eva_aud2txt}
\end{table}

\textbf{Image $\to$ Text:} 
We compare with four recent image captioning models: ATT~\cite{caption-att}, SAT~\cite{show-attend-and-tell}, RFNet~\cite{RFNet}, and UpDown (UD)~\cite{Anderson2017up-down}. The results are shown in Table~\ref{tab:eva_img2txt}. Note that our approach (\textbf{Ours}) uses a 3-layer CNN as the image encoder while all four baselines use deeper CNNs with pretraining/finetuning on extra datasets. Specifically, both ATT and SAT use the GoogleNet~\cite{Inception} which pretrained on ImageNet~\cite{ImageNet} as the image encoder. Our model, despite using a much shallower CNN, outperforms ATT and SAT by a large margin. The other two baselines use even more sophisticated image encoders: RFNet~\cite{RFNet} combines ResNet-101~\cite{ResNet}, DenseNet~\cite{DenseNet}, Inception-V3/V4/Resnet-V2~\cite{Inception}, all pretrained on ImageNet~\cite{ImageNet}. UpDown (UD)~\cite{Anderson2017up-down} uses a Faster R-CNN \cite{fasterRCNN} with Resnet-101~\cite{ResNet} pretrained on ImageNet~\cite{ImageNet} and finetuned on Visual Genome~\cite{VisualGenome} and COCO~\cite{MSCOCO}. For fair comparisons, we replace the 3-layer CNN with RFNet (\textbf{Ours w/ \cite{RFNet}}) and UD (\textbf{Ours w/ \cite{Anderson2017up-down}}). This improves performance compared to the baselines and shows the data efficiency of our approach: Because our model can handle multimodal data effectively, it can leverage external data sources \textit{even if} the modalities do not match. This helps our model learn more powerful multimodal representations from a large variety of data, which is not possible with the conventional bi-modal models.

\begin{table}[]
    \centering
    \small
    \begin{tabular}{lrr}
    \toprule
                       &  BLEU-1 / I2T  &  WER / S2T \\ \hline \hline
         Ours w/o \textbf{M}         &  65.2  &  6.99 \\ 
         Ours \textbf{M} $\to$ FC                 &  65.9  &  6.32 \\
         Ours w/o \textbf{T}         &  68.6  &  6.01 \\ 
         Ours w/o $L_{adv}$ &  69.8  &  5.87  \\ \hline
         Ours               &  \textbf{74.1}  &  \textbf{3.88} \\
         \bottomrule 
    \end{tabular}
    \vspace{0.2cm}
    \caption{\textbf{Ablation results} on image-to-text (I2T) and speech-to-text (S2T), evaluating contributions of different modules including the memory fusion module (M), the modality transformer (T), the adversarial loss ($L_{adv}$). \textbf{M} $\to$ FC means we replace \textbf{M} with two FC layers to match the number of parameters. }
    \label{tab:ablation}
    %\vspace{-3mm}
\end{table}

\textbf{Speech $\to$ Text:}
We compare our method with four ASR models, DeepSpeech2 \cite{DBLP:DeepSpeech2}, Seq2Seq \cite{speech-recog-seq2seq}, Policy Learning~\cite{Zhou2018ImprovingES}, and Gated Convnets~\cite{GatedConvnet}. All the models are trained end-to-end on the LibriSpeech corpus~\cite{LibriSpeech}. Particularly, similar to ours, the Seq2Seq model incorporates multi-head attention \cite{multi-head-attention} to attend to multiple locations of the encoded features. For a fair comparison, we fine-tune our model on the LibriSpeech corpus.%, replacing the vocabulary of EMT-4 with LibriSpeech. 

Table \ref{tab:eva_aud2txt} shows that our model outperforms the baselines on speech-to-text tasks. Our multimodal information bottleneck is trained on a larger variety of data, which helps them learn more powerful representations. Also, as the text modality comes from two unrelated datasets, the cross-modal reconstruction loss (Eqn.~\eqref{eq:recon}) enforces the model to solve more challenging optimization problems, which leads to improved results as seen in our experiments.

\textbf{Text $\to$ Speech:}
We compare with four text-to-speech (TTS) models: Tacotron~\cite{tacotron}, Tacotron 2~\cite{Tacotron2}, DeepVoice3~\cite{deepvoice3}, and GST~\cite{Tacotron-gst}. Tacotron~\cite{tacotron} is an RNN-CNN auto-regressive model trained only on reconstruction loss, while GST extends it by incorporating information bottleneck layers (which they call global style tokens). For both baselines we used the same Griffin-Lim algorithm~\cite{griffin-lim} as a vocoder. To evaluate the quality of the synthesized results quantitatively, we again use a pretrained ASR model based on WaveNet \cite{WaveNet} to compute the Word Error Rate (WER) for the samples synthesized by each model. Table \ref{tab:eva_aud2txt} shows that all three methods perform similarly. We believe one limiting factor is in the vocoder and expect to get better results with deep vocoders such as WaveNet~\cite{WaveNet}.

%--------------------------------------------------------
\textbf{Ablation Study.}
%\textbf{Contribution of different components in MIB.} 
We investigate the contribution of the modality transformer $\mathbf{T}$ and the memory fusion module $\mathbf{M}$, evaluating on image-to-text and speech-to-text tasks.

Table~\ref{tab:ablation} reports BLEU-1 scores for the image-to-text experiments and WER for the speech-to-text experiments. In both cross-modal generation tasks, the performance drops significantly when we remove the memory fusion module \textbf{M} (ours w/o $\mathbf{M}$). This suggests that the $\mathbf{M}$ plays the most significant role in modeling multimodal data. We also replace \textbf{M} with two FC layers that have a similar number of parameters as \textbf{M}. 
%The memory module has 41,281 parameters; replacing it with two FC layers with 80 units and 256 units, respectively, resulted in 41,296 parameters. 
This marginally improved the performance (B@1/I2T 65.2 vs. 65.9, WER/S2T 6.99 vs. 6.32). Our model still outperforms this baseline by a large margin (74.1 and 3.88). When we remove the modality transformer $\mathbf{T}$, we also see the performance drop significantly. This shows the importance of pushing modality-specific embeddings into a shared latent space; without this component, the $\mathbf{M}$ must deal with signals coming from three different modalities, which is a considerably more difficult task. We also test the contribution of the adversarial loss (Eqn.~\eqref{eq:loss_adv}) that we use to train $\mathbf{T}$. Without this loss term, the performance is similar to that in the setting without $\mathbf{T}$, which shows the adversarial loss plays a crucial role in training $\mathbf{T}$.
%\vspace{-5mm}

%%%%%%%%%%%%%%%%%%%%%%%%%%%%%%%%%%%%%%%%%%%%%%%%%%%%%%%%%
\section{Conclusion}
%%%%%%%%%%%%%%%%%%%%%%%%%%%%%%%%%%%%%%%%%%%%%%%%%%%%%%%%%
%\vspace{-2mm}
We propose a novel generative model for skip-modality generation. We demonstrate our approach on a challenging image-to-speech synthesis task where no paired data is available. Unlike conventional cross-modal generation, which relies on the availability of paired data, our model learns the correspondence between image and speech directly from two unrelated datasets, image-to-text and text-to-speech, using text as a shared modality. We show promising results on image-to-speech synthesis, as well as various cross-modal generation tasks, suggesting the model also benefits from increased data efficiency.
%\balance{}

{\small
\bibliographystyle{ieee_fullname}
\bibliography{main}
}

\newpage
\section*{Appendix}

%%%%%%%%%%%%%%%%%%%%%%%%%%%%%%%%%%%%%%%%%%%%%%%%%%%%%%%%%%%%%%%%%%%%%%%%%%%%%%%%%%%%%%%%%%%
\section{Network Architectures and Parameter Settings}
%%%%%%%%%%%%%%%%%%%%%%%%%%%%%%%%%%%%%%%%%%%%%%%%%%%%%%%%%%%%%%%%%%%%%%%%%%%%%%%%%%%%%%%%%%%

We provide implementation details of our model with the parameter settings used in our experiments. We encourage the readers to refer to Figure 3 and Figure 4 of our main paper when reading this section. We use the following notations to refer to commonly used computation blocks in the neural networks: 
Conv1D(\#channels, kernel\_size, stride\_size), Conv2D(\#channels, kernel\_size, stride\_size), FC(\#units), GRU(\#units). $\oplus(\cdot)_{res}$ refers to the residual connection. We use the superscript $j$ to refer to the modalities $j \in \{img, txt, spch\}$.

\subsection{Encoders (Figure 3 (left) in the main paper)}
\begin{itemize}
    \item \textbf{Image encoder:} $\mathbf{x}^{img}$ $\to$ Conv2D(64, 4, 2) $\to$ BN $\to$ ReLU $\to$ Conv2D(128, 4, 2) $\to$ BN $\to$ ReLU $\to$ Conv2D(256, 4, 2) $\to$ BN $\to$ ReLU $\to$ Conv2D(512, 4, 2) $\to$ BN $\to$ ReLU $\to$ MaxPool $\to$ $\mathbf{e}^{img}$
    
    \item \textbf{Text encoder:} $\mathbf{x}^{txt}$ $\to$ LookupTable(66, 128) $\to$ FC(256) $\to$ ReLU $\to$ Dropout(0.5) $\to$ FC(128) $\to$ ReLU $\to$ Dropout(0.5) $\to$ CBHG~\cite{tacotron} $\to$ AvgPool $\to$ FC(512) $\to$ tanh $\to$ $\mathbf{e}^{txt} \in \mathbb{R}^{512}$
    
    \item\textbf{Speech encoder:} $\mathbf{x}^{spch}$ $\to$ Conv2D(32, 3, 2) $\to$ BN $\to$ ReLU $\to$  Conv2D(64, 3, 2) $\to$ BN $\to$ ReLU $\to$ GRU(256) $\to$ FC(512) $\to$ tanh $\to$ $\mathbf{e}^{spch}$
\end{itemize}

\subsection{Multimodal Information Bottleneck (Figure 4 in the main paper)}
\begin{itemize}
    \item\textbf{Modality transformer:} $\mathbf{e}^{j}$ $\to$ FC(256) $\to$ ReLU $\to$ $\oplus \Big($ Conv1D(256, 1, 1) $\to$ ReLU $\to$ BN $\to$ Conv1D(256, 1, 1) $\to$ ReLU $\to$ BN) $\Big)_{res}$ $\to$ tanh $\to$ $\mathbf{z}^{j}$

    \item\textbf{Memory fusion module:} 
        \begin{enumerate}
            \item Define: Memory $M \in \mathbb{R}^{n_k \times d_k/n_{heads}}$, where $n_k=128, d_k=256, n_{heads}=4$
            
            \item Query $\mathbf{q}^{j}  \mathbf{z}^{j}$, Key $\mathbf{k}$ Conv1D(256, 1, 1)(tanh($M$)), Value $\mathbf{v}$ tanh($M$)
            
            \item $(\mathbf{q}_{h}^{j}, \mathbf{k}_{h}, \mathbf{v}_{h}) \mbox{SplitHeads}(\mathbf{q}^{j}, \mathbf{k}, \mathbf{v})$, $h=1, \cdots, n_{heads}$
            
            % _mlp_attention
            %\item $\alpha_{h}^{j} \mapsfrom $ SoftMax$\left( \frac{\mathbf{v}_{rand}}{\|\mathbf{v}_{rand}\|} \times \mbox{tanh} \left( \mbox{FC}(d_k)(\mathbf{q}_{h}^{j}) + \mbox{FC}(d_k)(\mathbf{k}_{h}) + bias \right) / \sqrt{d_k} \right)$, $h=1, \cdots, n_{heads}$
            
            % _dot_attention
            \item $\alpha_{h}^{j} \mbox{SoftMax} \left( \mathbf{q}_{h}^{j} \mathbf{k}_{h} / \sqrt{d_k} \right), h=1, \cdots, n_{heads}$
            
            \item $\mathbf{u}_{h}^{j} \alpha_{h}^{j} \times \mathbf{v}_{h}$, $h=1, \cdots, n_{heads}$
            
            \item $\mathbf{u}^{j}  $ ConcatHeads$(\mathbf{u}_{h}^{j})$
            
            %\item SplitHeads(qs, ks, vs) $\to$ MultiHead \{normalize(vs) $\times$ tanh(FC(qs) + FC(ks) + bias) $\to$ FC(256) $\to$ SoftMax \} $\to$ CombineHeads $\to$ WeigtSum
        \end{enumerate}
         
\end{itemize}

\subsection{Decoders (Figure 3 (right) in the main paper)}
\begin{itemize}
    \item \textbf{Image decoder:} $\mathbf{u}^{txt}$ $\to$ Conv2D$^{\intercal}$(32, 4, 2) $\to$ BN $\to$ ReLU $\to$ Conv2D$^{\intercal}$(16, 4, 2) $\to$ BN $\to$ ReLU $\to$ Conv2D$^{\intercal}$(8, 4, 2) $\to$ BN $\to$ ReLU $\to$ Conv2D$^{\intercal}$(8, 4, 2) $\to$ tanh $\to$ $\mathbf{y}^{img}$
    
    \item \textbf{Text decoder:} $\mathbf{u}^{img}$ $\to$ Dropout(LSTM(128), 0.3) $\to$ Dropout(LSTM(128), 0.3) $\to$ FC($n_{symbols}$) $\to$ SoftMax $\to$ $\mathbf{y}^{txt}$
    
    \item \textbf{Speech decoder:} $\mathbf{u}^{txt}$ $\to$ AttentionRNN(GRU 256) $\to$ DecoderRNN6(Dropout(LSTM(256), 0.3) $\to$ Dropout(LSTM(256), 0.3)) $\to$ reshape $\to$ $\mathbf{y}^{mel-spectrogram}$ $\to$ CBHG~\cite{tacotron} (80 mels) $\to$ FC(1025) $\to$ Griffin-Lim ($\mathbf{y}^{linear}$) $\to$ $\mathbf{y}^{speech}$
    
\end{itemize}

%%%%%%%%%%%%%%%%%%%%%%%%%%%%%%%%%%%%%%%%%%%%%%%%%%%%%%%%%%%%%%%%%%%%%%%%%%%%%%%%%%%%%%%%%%%
\section{Skip-Modal Synthesis Results}
%%%%%%%%%%%%%%%%%%%%%%%%%%%%%%%%%%%%%%%%%%%%%%%%%%%%%%%%%%%%%%%%%%%%%%%%%%%%%%%%%%%%%%%%%%%
Figure~\ref{fig:sup_example1} shows additional image-to-speech synthesis results; we manually transcribed the synthesized audio outputs for the purpose of presentation. Consistent with the qualitative results reported in the main paper (Figure 5), we see that our approach produces more detailed descriptions and has a larger vocabulary than the baseline. We encourage the readers to visit our anonymized website and listen to the audio samples: \url{https://bit.ly/2U7741S}

\begin{figure*}
    \centering
    \includegraphics[width=\linewidth]{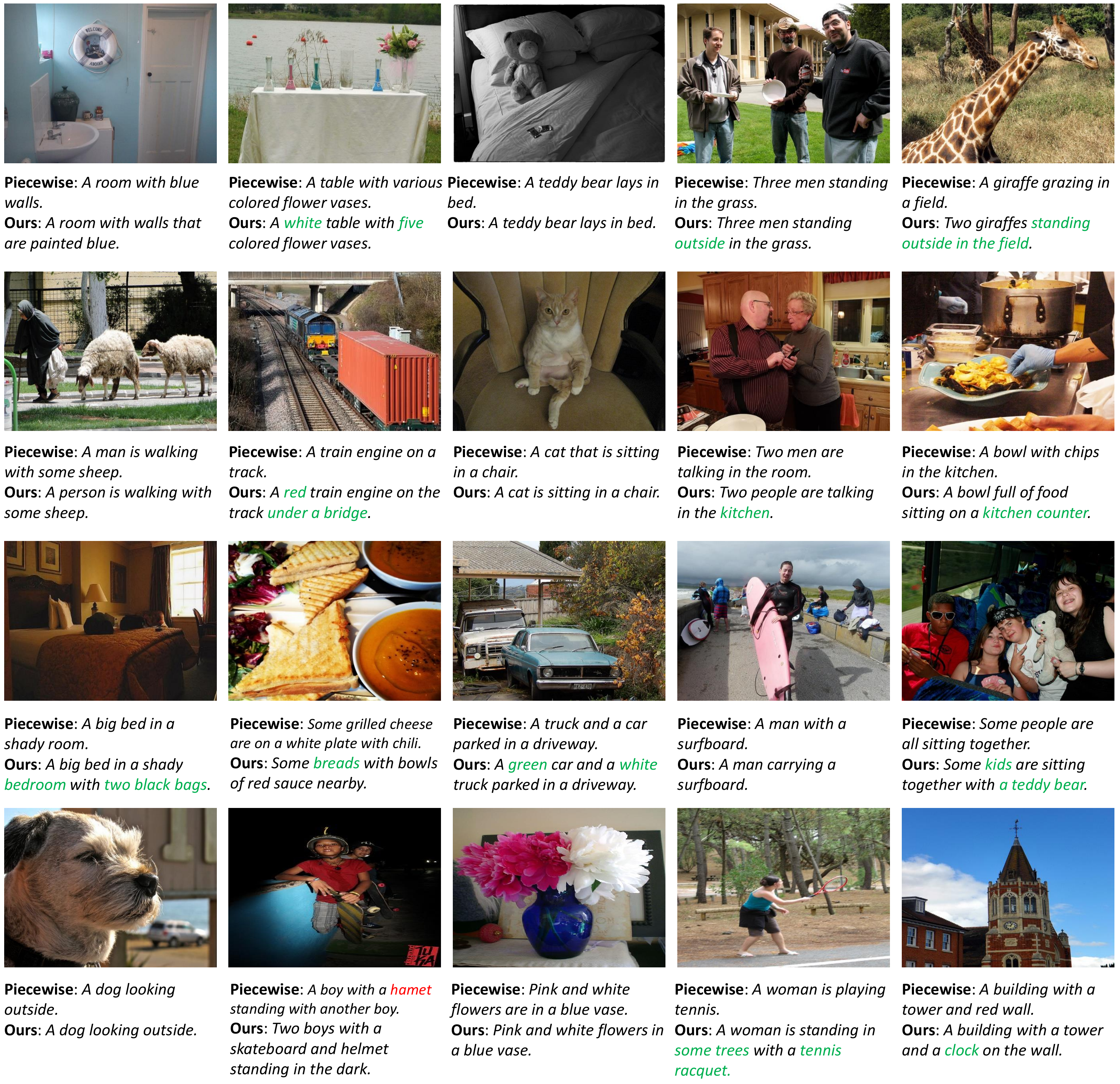}
    \caption{Image-to-speech synthesis results. \textcolor{mygreen}{Green}: {Fine-grained and correct instances synthesized by our model.} \textcolor{myred}{Red}: {incorrect pronunciation synthesized by the piecewise model}. Audio samples are available at \url{https://bit.ly/2U7741S}}
    \label{fig:sup_example1}
\end{figure*}

%%%%%%%%%%%%%%%%%%%%%%%%%%%%%%%%%%%%%%%%%%%%%%%%%%%%%%%%%%%%%%%%%%%%%%%%%%%%%%%%%%%%%%%%%%%
\section{Cross-Modal Retrieval Results}
Besides the synthesis tasks, another way to evaluate the performance of our model is via cross-modal retrieval. In this section, we show qualitative results of cross-modal retrieval where we use an instance from either dataset and find the most similar instances from different modalities from both datasets. Specifically, we compute $\mathbf{u}^{j}$ from all instances in the test splits of both datasets, and compute the cosine similarity between any pair of cross-modal instances.

\begin{figure*}
    \centering
    \includegraphics[width=\linewidth]{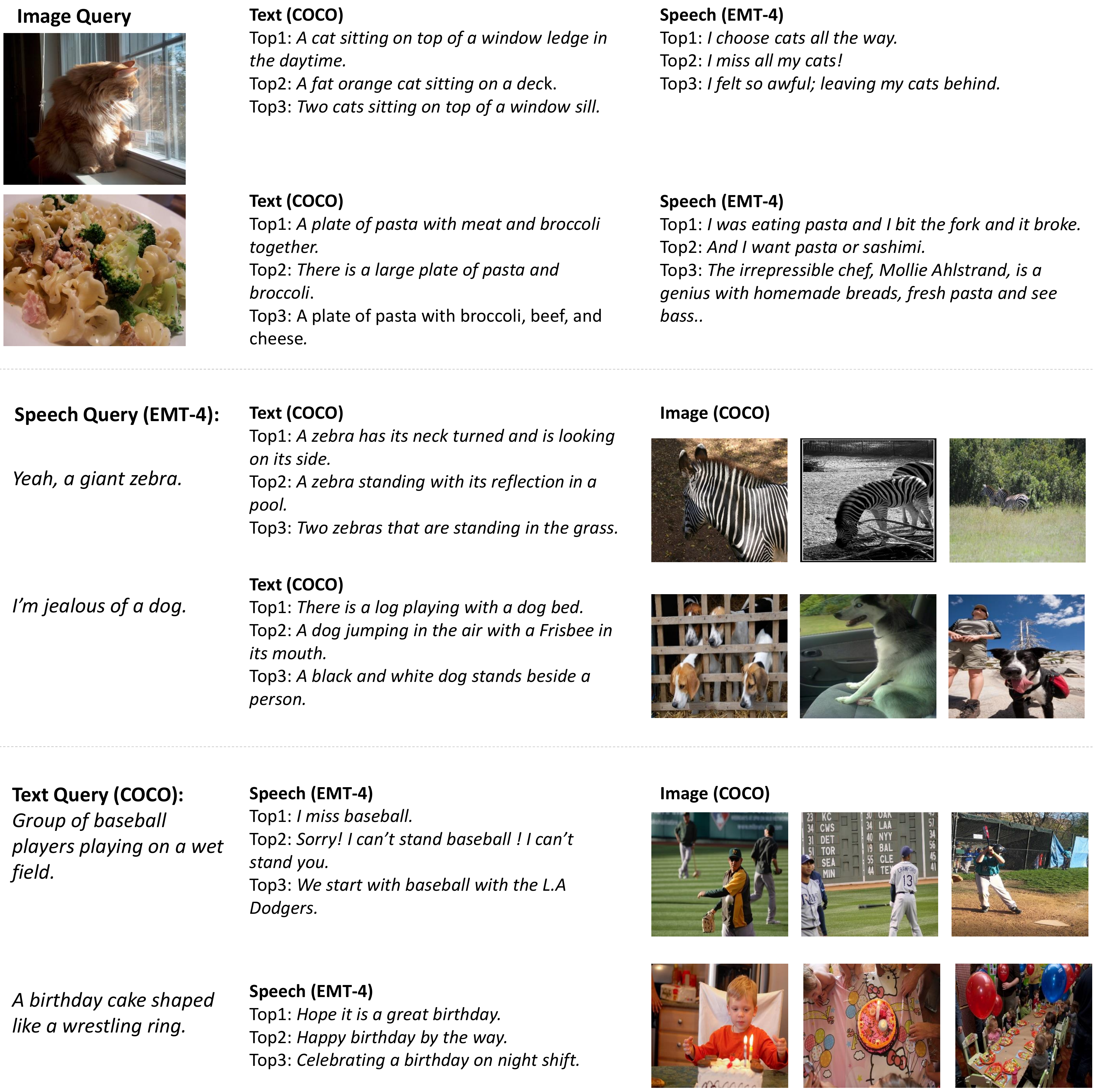}
    \caption{Cross-Modal retrieval results. The first column shows queries from each modality. The second and third columns show the top-3 retrieval results from the other two modalities. Audio samples are available at \url{https://bit.ly/2U7741S}}
    \label{fig:sup_retrieval}
\end{figure*}

Figure~\ref{fig:sup_retrieval} shows the top 3 retrieved results in different combinations of modalities. We can see that the retrieved results are very related to the query at the object level, e.g., ``dog'' and ``zebra'' in the first and the second rows, while on the other two rows the results are related to the query at the scene/context level, e.g., ``baseball game'' and ``birthday party''. It is particularly interesting to see that the results are reasonable even for the cross-dataset retrieval settings (using an image from COCO to retrieve audio/speech instances from EMT-4). This suggests the representations extracted using our model are not very sensitive to the dataset and the modalities involved.

%To validate our proposed model has learned alignment across different modalities, we perform a cross modal retrieval experiment. Given a query input from one modality, we quantitative visualize how well the learned representation retrieve its corresponding pair other modalities. Specifically, we input a query from one modality into our network, and take the output from the Memory Fusion Module as the latent representation. We then rank the samples from other modalities by nearest cosine similarity. The results are shown in Figure \ref{fig:sup_retrieval}. As we can see that, the results are highly related to the query in instance level, e.g. 'dog' and 'zebra' in first and second row. The results on the third row show that, the learned representations also aligned on semantic level which depends on not only the instance but also a global context, e.g. 'birthday party'. 

%%%%%%%%%%%%%%%%%%%%%%%%%%%%%%%%%%%%%%%%%%%%%%%%%%%%%%%%%%%%%%%%%%%%%%%%%%%%%%%%%%%%%%%%%%%

%%%%%%%%%%%%%%%%%%%%%%%%%%%%%%%%%%%%%%%%%%%%%%%%%%%%%%%%%%%%%%%%%%%%%%%%%%%%%%%%%%%%%%%%%%%
\section{Additional Ablation Experiments}
%%%%%%%%%%%%%%%%%%%%%%%%%%%%%%%%%%%%%%%%%%%%%%%%%%%%%%%%%%%%%%%%%%%%%%%%%%%%%%%%%%%%%%%%%%%

%------------------------------------------------------------------------------------------
\subsection{Different Batch Sampling Strategies}
%------------------------------------------------------------------------------------------
As we trained our model on a combination of two datasets, there comes two ways to perform mini-batch training: one that samples instances from only one dataset and alternates between the two (alternate); and another that always samples instances from both datasets (mixing). We compare these two batch sampling strategies in this section. Specifically, in the first setting (alternative) we sample eight instances from either the COCO or EMT-4 dataset, while in the second setting (mixing) we sample four instances from COCO and the other four from EMT-4. We evaluate this on image-to-text (I2T), speech-to-text (S2T), and text-to-speech (T2S) synthesis tasks, reporting BLEU-1 for I2T and the word error rate (WER) for the other two.

\begin{table}[]
    \centering
    % \begin{tabular}{|c||c|c|c|}
    \begin{tabular}{p{2cm}p{0.8cm}p{0.8cm}p{0.8cm}}
    \hline 
    Batch Sampling Strategy & B@1 (I2T) & WER (S2T) & WER (T2S)\\ \hline  
    Alternative     & 74.1 & 3.88 & 10.5 \\ \hline 
    Mixing          & 74.5 & 3.76 & 10.5 \\ \hline
    \end{tabular}
    \caption{Evaluation of different batch sampling strategies. I2T: image-to-text, S2T: speech-to-text, T2S: text-to-speech.}
    \label{tab:batch_sampling}
\end{table}
Table~\ref{tab:batch_sampling} shows that the performance improves when we use the mixed batch sampling strategy. The improvement is especially pronounced for the text-sensitive tasks; on image-to-text synthesis the BLEU-1 is improved from 74.1 to 74.5, and on speech-to-text synthesis the WER is reduced from 3.88 to 3.76. We did not find significant differences in the text-to-speech synthesis task.%While for text-to-images and text-to-speech tasks, we don't see obvious differences. 

%------------------------------------------------------------------------------------------
\subsection{Sensitivity Analysis of Memory Fusion Module with parameters $n_k$ and $d_k$}
%------------------------------------------------------------------------------------------
As we showed in Table 4 (ablation results) in the main paper, the memory fusion module plays an important role in our model; the performance drops most significantly when we bypass this module. It extracts compact, modality-agnostic representations of the multimodal inputs following the information bottleneck principle~\cite{tishby1999information}, using the shared external memory $M$ to ``bottleneck'' any redundant and modality-specific information from leaking into the output representation. To better understand the behavior of this module, we analyze the sensitivity of the module to two hyper-parameters: the number of basis vectors ($n_k$) and the size of each basis vector ($d_k$) inside the external memory variable $M$. We evaluate this on image-to-text generation (i.e., image captioning) and report the results using BLEU-1 as our metric.

\begin{table}
    \centering
    \begin{tabular}{|c|c|c|c|}
    \hline 
     $n_k$ (fix $d_k=256$)   & 10    & 128  & 256 \\ \hline 
       BLEU-1                               & 62.5  & 74.1 & 73.9 \\ \hline \hline  
     $d_k$ (fix $n_k=128$)     & 64   & 128   & 256  \\ \hline
       BLEU-1                               & 49.3 & 70.2  & 74.1 \\ \hline 
    \end{tabular}
    \caption{Sensitivity of the memory fusion module. $n_k:$ the number of basis vectors, $d_k$: the size of each basis vector.}
    \label{tab:sup_parameter}
\end{table}
Table~\ref{tab:sup_parameter} shows our model is more sensitive to the dimension of each basic vector $d_k$ than the number of basis vectors $n_k$; it achieves a significantly lower performance with $d_k=64$ compared to any other combination of the two parameter values. The performance improves as we increase $d_k$, achieving the best performance when $d_k=256$; we did not evaluate beyond $d_k=256$ due to the limitations on the GPU memory. As for the number of basis vectors $n_k$, we can see the performance is low when there are only a few of them ($n_k=10$). This shows we need a large number of basis vectors to capture the variety of information contained in multimodal data. We found that the performance is relatively stable when $n_k$ is greater than 128.

\end{document}